\documentclass[preprint,12pt,authoryear]{elsarticle}

\usepackage{microtype}
\usepackage{graphicx}
\usepackage{subfigure}
\usepackage{booktabs} 

\usepackage{verbatim}
\usepackage{subcaption}
\usepackage{pgfplots}
\pgfplotsset{compat=newest}

\usepackage{array}     
\usepackage{adjustbox}
\usepackage{multirow}  
\usepackage{hyperref}

\usepackage{amsmath}
\usepackage{amssymb}
\usepackage{mathtools}
\usepackage{amsthm}
\usepackage{amsfonts}
\usepackage{algorithmic}
\usepackage[capitalize,noabbrev]{cleveref}

\theoremstyle{plain}
\newtheorem{theorem}{Theorem}[section]

\newtheorem{lemma}[theorem]{Lemma}

\theoremstyle{definition}

\theoremstyle{remark}

\usepackage[textsize=tiny]{todonotes}
\usepackage{amssymb}
\usepackage{amsmath}
\usepackage{algorithm}

\newcommand{\norm}[1]{\left\lVert #1\right\rVert_2}

\journal{Expert Systems With Applications}

\begin{document}

\begin{frontmatter}



\title{FedRP: A Communication-Efficient Approach for Differentially Private Federated Learning Using Random Projection} 

\author[label1]{Sina Najafi}
\author[label1]{Mostafa Tavassolipour} 
\author[label1]{Mohammad Hasan Narimani}
\affiliation[label1]{organization={School of Electrical and Computer Engineering},
            addressline={ College of Engineering},
            city={University of Tehran},
            country={Iran}}

\begin{abstract}
Federated learning (FL) enables collaborative model training without centralizing data, but exchanging high-dimensional updates can expose sensitive information and incur substantial communication costs. We present FedRP, a communication-efficient method combining Gaussian random projection with consensus optimization based on the alternating direction method of multipliers (ADMM). In each round, clients project their model parameters into an $m$-dimensional space using a shared random matrix hidden from the server, which aggregates only compressed representations. We establish a high-probability guarantee linking projected-space consensus to proximity among client models and derive an $(\epsilon,\delta)$-differential privacy guarantee for each release under bounded $\ell_2$-sensitivity and a positive lower bound on parameter norms. Withholding full parameter vectors and the projection matrix also limits information available to common reconstruction attacks. Experiments on MNIST and CIFAR-10 with LeNet-5 and a custom convolutional network show that FedRP achieves accuracy comparable to FedAvg and consistently exceeds noise-perturbed, differentially private FedAvg. Because clients transmit $m$ rather than $n$ values per round, FedRP reduces communication by orders of magnitude when $m\ll n$. The results demonstrate a favorable trade-off among accuracy, privacy, and communication efficiency. Code is available at https://github.com/mhnarimani/FedRP.
\end{abstract}




\end{frontmatter}




\section{Introduction}

In the past decade, advancements in machine learning have fundamentally reshaped how we utilize data, impacting daily life through applications from smartphones to autonomous vehicles. Globally, a vast and ever-growing number of decentralized devices incessantly generate complex data. While machine learning models were historically trained on large, centralized datasets requiring robust hardware and storage, the exponential growth in data volume has made such approaches increasingly impractical. This has led to the rise of distributed learning, a paradigm where processing occurs concurrently across many devices, which collaborate by exchanging information to build a comprehensive and efficient global model \cite{tavassolipour2017learninggaussianprocessesdistributed}. Among the most significant innovations in this area is Federated Learning (FL), a framework that enables collaborative model training while upholding privacy principles, making it a critical technology for a data-conscious future and a foundational element in the advancement of intelligent systems \cite{reee1,9599369}.

Federated Learning is a machine learning paradigm that reverses the traditional training process; instead of bringing data to a central model, the model is brought to the data. In the FL framework, each client device independently trains a model using its own local data. After a training cycle, the resulting model updates—such as parameters or gradients—are transmitted to a central server. The server then aggregates these updates to enhance a shared global model, and this iterative process continues until the model achieves its targeted accuracy \cite{mcmahan2017federated}. The principal advantage of this approach is a significant enhancement in user privacy, as raw data is never transferred from the device to a central server. This process is applied across diverse domains, including improving predictive accuracy on smartphones \cite{hard2019federatedlearningmobilekeyboard,ahmed2024ondevicefederatedlearningsmartphones}, safeguarding privacy in Internet of Things (IoT) systems \cite{YAACOUB2023155,elzemity2024privacythreatscountermeasuresfederated,inproceedings}, and enabling analysis of sensitive medical data without sharing raw patient information \cite{guan2024federatedlearningmedicalimage,dogggg,engproc2023059230,fdb74649b0744e45afc3d2fe3232c809,articleslkj}.

Federated learning problems are generally categorized into two main scenarios: cross-device and cross-silo. Cross-device FL involves a massive number of clients, such as mobile phones or IoT sensors, which are often unreliable, have limited computational resources, and may have intermittent connectivity. In this setting, only a small fraction of clients typically participates in any given training round. The data is often highly personalized and therefore not independent and identically distributed (non-IID), reflecting unique user behaviors and preferences, as exemplified in predicting words on a smartphone keyboard \cite{xia2023crossdevicefederatedlearningmobile,chen2023fsrealrealworldcrossdevicefederated,9425266}.

In contrast, cross-silo FL involves a much smaller number of participants, which are typically organizations like hospitals, banks, or research institutions. These ``silos'' are stable, reliable, and possess significant computational resources. Each silo holds a large, often more structured, dataset. Due to their reliability and powerful infrastructure, it is common for all or most clients to participate in every training round \cite{huang2022crosssilofederatedlearningchallenges,gorbett2024crosssilofederatedlearningdivergent}. While data may still be heterogeneous across silos (e.g., reflecting different patient demographics at various hospitals), the environment is significantly more controlled and the number of participants is manageable compared to the cross-device setting.

Moreover, FL contends with statistical heterogeneity, contrasting with traditional methods where data is uniformly distributed across devices. In FL, local data frequently exhibits heterogeneity, as exemplified in predicting words on a smartphone keyboard, where each user’s data may vary. This heterogeneity can present challenges in model convergence \cite{hard2019federatedlearningmobilekeyboard,aa,dad}.

Despite its merits, Federated Learning (FL) faces several critical challenges that distinguish it from other distributed learning approaches and must be addressed to enable its widespread adoption. One of the most prominent issues is the communication cost. Modern deep learning models often comprise hundreds of millions of parameters, resulting in extremely large model updates that must be transmitted by each client. In networks involving millions of devices, the cumulative bandwidth required becomes a significant bottleneck, highlighting the need for efficient compression techniques and communication protocols \cite{Asad2023LimitationsAF,Hayyolalam2023CommunicationEC,Wen2022ASO,tavassolipour2018structurelearningsparseggms,Tavassolipour_2019}.

Another major concern is privacy preservation. While FL prevents raw data from leaving user devices, the transmitted model updates are not inherently secure. Research has demonstrated that adversaries with access to shared gradients can perform reconstruction attacks, extracting sensitive personal information from the underlying data. This threatens the foundational privacy guarantees of FL and necessitates the integration of robust privacy-enhancing techniques into the training process \cite{computers13110277,aldaghri2023federatedlearningheterogeneousdifferential,wei2025fedcaprivacyprivacypreservingheterogeneousfederated,10d3679013}.

Finally, statistical heterogeneity poses a unique challenge. In FL, data across devices is typically non-independent and identically distributed (non-IID), leading to divergent local training objectives. This heterogeneity can cause inconsistencies in model updates and hinder the convergence of the global model, making it difficult to train an effective and generalizable model across all participants \cite{hard2019federatedlearningmobilekeyboard,aa,dad}.

This paper introduces a novel algorithm, FedRP, designed to confront the dual challenges of privacy preservation and communication cost reduction. Our approach integrates random projection techniques with the Alternating Direction Method of Multipliers (ADMM) optimization framework \cite{boyd}. The core idea is to have clients project their high-dimensional model parameters into a secure, low-dimensional space using a shared random matrix before transmission. This projection creates a one-way information bottleneck; it is computationally infeasible for an attacker or the server to reconstruct the original model parameters from the compressed vector without access to the secret matrix, which is generated securely by the clients in each round (The server does not have access to this matrix; only the clients have access to it). The ADMM algorithm then provides a powerful framework for driving the clients to a consensus in this projected space. We demonstrate theoretically that FedRP provides a strong $(\epsilon, \delta)$-differential privacy guarantee and is resilient to data reconstruction attacks. Furthermore, our extensive experiments show that this method achieves a trifecta of benefits: it maintains high model accuracy comparable to conventional algorithms like FedAvg, drastically reduces communication costs, and provides quantifiable privacy enhancements. 

In this work, we focus on a federated learning scenario where all clients participate in each iteration of the training process. Such an assumption is standard and highly reasonable in the cross-silo FL setting, where a limited number of reliable and resourceful institutions collaborate. This focus allows us to rigorously analyze the core properties of our algorithm without the added complexity of client selection and dropouts. While FedRP algorithm could be extended to the more dynamic cross-device case where only a subset of clients participates in each round, that extension is beyond the scope of this article. The main contributions of this paper are organized as follows:

\begin{enumerate}
	\item We introduce FedRP, a cross-silo FL framework that combines Gaussian random projection with ADMM. Clients transmit only low-dimensional projected parameters, while the shared projection matrix remains unavailable to the server.
	
	\item We prove that consensus among projected client models implies proximity among their original parameters with high probability. Under bounded $\ell_2$-sensitivity and a positive lower bound on the parameter norm, we also derive an explicit $(\epsilon,\delta)$-differential privacy guarantee for each release, discuss multi-round composition, and analyze protection against common reconstruction attacks.
	
	\item Experiments on MNIST and CIFAR-10 show that FedRP achieves accuracy close to FedAvg and consistently outperforms noise-perturbed FedAvg with differential privacy. We also evaluate the effect of the projection dimension on convergence and accuracy.
	
	\item We show that transmitting an $m$-dimensional projection instead of an $n$-dimensional parameter vector reduces per-round communication by orders of magnitude when $m\ll n$, providing a favorable utility--privacy--communication trade-off.
\end{enumerate}

\section{Related Work}

Federated learning (FL) protects privacy by avoiding raw data sharing, but exchanging model gradients and parameters can still expose sensitive information. Attack methods like Data Leakage from Gradient (DLG) and Deep Leakage from Model (DLM) exploit gradients and consecutive model snapshots to reconstruct private training data, including class labels and original samples \cite{DLG,dlm,10646724,biswas2024bayescapacitymeasurereconstruction,10793076,10854547,10945712}. While techniques like GradInversion have improved batch data recovery, challenges remain, particularly for larger batches \cite{yin2021gradientsimagebatchrecovery,11009149}. These risks highlight the ongoing challenge of preventing data reconstruction attacks and ensuring robust privacy in FL.

Privacy preservation in federated learning (FL) is challenging due to data heterogeneity and vulnerabilities in model updates. Various techniques, including Homomorphic Encryption (HE) \cite{z90,z91,z92}, Secure Multiparty Computation (SMC) \cite{z93,z94}, and Differential Privacy (DP) \cite{z95,dpbook,qqq3}, aim to mitigate these risks. HE enables computations on encrypted data, preventing information leakage during gradient exchange, but suffers from high computational costs, especially in its fully homomorphic form. Studies have proposed hybrid models, such as integrating HE with DP, to balance security and efficiency \cite{z31}. However, practical implementation remains difficult due to processing overhead and complexity. Recent work has begun to address these limitations. For instance, a verifiable privacy-preserving federated learning scheme based on homomorphic proxy re-encryption splits ciphertext components across two semi-trusted servers, preventing any single server from recovering individual gradients while providing verifiability through digital signatures and linear homomorphic hashing \cite{SONG2026122875}. In a complementary direction, multi-key homomorphic encryption combined with Shamir-style linear secret sharing, such as CDKS-LSS, removes the need for a trusted third party, supports threshold-based aggregation under client dropout, and offers post-quantum security guarantees for federated training and fine-tuning of high-capacity models \cite{WANG2026123619}.

SMC enhances privacy by allowing multiple parties to collaboratively compute functions without revealing individual data, making it useful for securely aggregating model updates in FL. However, it introduces significant computational and communication overhead, limiting scalability. Hybrid approaches, like combining SMC with DP \cite{vs}, have shown promise in reducing information leakage and improving security against data extraction attacks. Despite these advancements, challenges such as coordination among clients and efficiency trade-offs persist.

Differential privacy (DP) is a key approach to ensuring privacy in federated learning (FL) by introducing noise to gradients, model parameters, or objective functions. Global DP applies noise at the dataset level, balancing privacy and performance \cite{azv,wei2020federated,9069945,9714350,qqq3}. In contrast, local DP secures client-server interactions with stronger privacy guarantees but higher noise \cite{z27,z107}.

To counter data leak attacks, techniques like noisy gradients, gradient compression, and synthetic data generation have been proposed. For example, fake gradients have been introduced, and local perturbation with global update compensation has been used to maintain accuracy \cite{z114}. Adaptive DP models have also been developed to dynamically allocate privacy budgets and minimize performance loss \cite{fu2022adapdpfldifferentiallyprivate}. While DP methods enhance privacy, they often come at the cost of model performance, posing an ongoing challenge. Recent work has also highlighted the intrinsic inter-linkages between privacy, security, and fairness in federated learning. In particular, \cite{qqq4} provide a comprehensive survey revealing the trade-offs among these three dimensions and propose fairness as a key mediator between privacy and security. 
Their analysis suggests that improvements in one aspect can significantly influence the others, offering new perspectives for designing more robust and balanced federated learning systems.

Recently, compression-based techniques have shown that careful sketching and projection can simultaneously enhance privacy, reduce communication, and preserve model accuracy in differentially private federated learning (DP-FL). Secure aggregation schemes using sparse random projections achieve significant bandwidth savings without sacrificing utility \cite{chen2022fundamentalpricesecureaggregation}. Count-sketch methods, such as FedSKETCH and its heterogeneous variant, transmit compressed, unbiased gradients under differential privacy, ensuring convergence for a wide range of objectives \cite{FedSKETCH}. A related line of research demonstrates that simple random compression is sufficient to amplify privacy in both central and shuffled models, substantially reducing communication for mean and frequency estimation tasks \cite{chen2023privacyamplificationcompressionachieving}. Iterative sketching with fast Johnson-Lindenstrauss transforms further lowers per-round costs while maintaining standard descent convergence under formal DP analyses \cite{song2022iterative}. Finally, FetchSGD combines count sketches with server-side momentum and error feedback to deliver robust convergence on non-i.i.d. data with sparse client participation, outperforming FedAvg in practical federated learning scenarios \cite{FetchSGD}.
The overall structure of FedRP algorithm is very similar to previous works that utilized sketching and projection techniques, but there is a fundamental difference between these methods and FedRP algorithm. Unlike similar methods such as FedSKETCH, FedRP algorithm does not require reconstructing the dimension-reduced vectors back to their original dimensions. This difference results in FedRP algorithm being both faster and having lower error.

\section{Methodology}

Federated learning was initially considered to be privacy-preserving; however, studies have shown that adversaries can reconstruct original data from shared gradients \cite{dlm,DLG}. To address this, techniques like noise addition have been used, though they often compromise model accuracy. Balancing privacy, accuracy, and communication efficiency remains a major challenge. This work proposes a method that enhances data privacy while preserving model performance and reducing communication overhead.

\subsection{Consensus ADMM Optimization}\label{admmmmm}

The consensus problem is one of the common problems in distributed optimization, where multiple agents aim to achieve an optimal and shared value for a variable \cite{boyd}. However, each agent can have different objectives compared to other agents. In FL, clients train a shared model under the coordination of a central server without sharing their local data. In practice, the goal of Federated Learning is to achieve consensus among clients on a shared global model. Therefore, Federated Learning can be framed within the context of the consensus problem. Therefore, according to the consensus problem framework, each client should optimize its model based on its local data while striving to align its model parameters with a common global value.

Consider a set of \( K \) clients, each with its own local dataset \( \mathcal{D}_i \) and the local model parameter \( \mathbf{w}_i \) for \( i = 1, 2, \ldots, K \). The objective is to find a global model parameter vector \( \bar{\mathbf{z}} \) that minimizes the sum of the local objective functions \( f_i(\bar{\mathbf{z}}) \) defined by the local datasets. This can be formulated as
\begin{equation}
\min_{\mathbf{w}_1, \mathbf{w}_2, \ldots, \mathbf{w}_K} \sum_{i=1}^{K} f_i(\mathbf{w}_i)
\end{equation}
subject to the constraint
\begin{equation}
\mathbf{w}_i = \bar{\mathbf{z}} \quad \forall i = 1, ..., K.
\end{equation}

In this problem, we assume that the distribution of client data is identical and that all samples within the clients are distributed independently and identically (IID). To solve this problem, various optimization techniques can be applied, including the Alternating Direction Method of Multipliers (ADMM), which is particularly effective in distributed settings as it decomposes the problem into sub-problems that can be solved in parallel by each client \cite{qqq1,qqq2}. The augmented Lagrangian for this problem is:

\begin{equation}
\label{eq3}
L_\rho(\{\mathbf{w}_i\}, \bar{\mathbf{z}}, \{\mathbf{y}_i\}) = 
\sum_{i=1}^K \left( f_i(\mathbf{w}_i) + \mathbf{y}_i^T (\mathbf{w}_i - \bar{\mathbf{z}}) + \frac{\rho}{2} \| \mathbf{w}_i - \bar{\mathbf{z}} \|_2^2 \right)
\end{equation}

where \(\{\mathbf{y}_i\}\) are scaled dual variables and \(\rho > 0\) is a penalty parameter. The second term corresponds to the Lagrange multipliers used to enforce the constraint \( \mathbf{w}_i = \bar{\mathbf{z}}\), while the third term penalizes the deviation of local parameters from the global model, helping to keep them close during training. The ADMM algorithm proceeds by iteratively updating the local variables \(\{\mathbf{w}_i\}\), the global variable \(\bar{\mathbf{z}}\), and the dual variables \(\{\mathbf{y}_i\}\) as follows:

1.Update \(\mathbf{w}_i\) (Local Update):

\begin{equation}\label{eq4}
\mathbf{w}_i^{t+1}\ = \arg \min_{\mathbf{w}_i} \left( f_i(\mathbf{w}_i) + (\mathbf{y}_i^t)^T(\mathbf{w}_i - \bar{\mathbf{z}}^t) + \frac{\rho}{2} \| \mathbf{w}_i - \bar{\mathbf{z}}^t \|_2^2 \right)
\end{equation}

2. Update \(\bar{\mathbf{z}}\) (Global Update):

\begin{equation}\label{eq5}
\bar{\mathbf{z}}^{t+1}\  = \frac{1}{K} \sum_{i=1}^K \left( \mathbf{w}_i^{t+1} + \frac{1}{\rho} \mathbf{y}_i^t \right)
\end{equation}

3. Update \(\mathbf{y}_i\) (Dual Variable Update):

\begin{equation}\label{eq6}
\mathbf{y}_i^{t+1}\ = \mathbf{y}_i^t + \rho \left( \mathbf{w}_i^{t+1} - \bar{\mathbf{z}}^{t+1} \right)
\end{equation}
Given \(\sum_{i=1}^K \mathbf{y}_i^t = 0\), equation 5 can be simplified as follows:
\begin{equation}\label{eq7}
\bar{\mathbf{z}}^{t+1}\  = \frac{1}{K} \sum_{i=1}^K  \mathbf{w}_i^{t+1} 
\end{equation}
By performing the above steps sufficiently, the global variable \( \bar{\mathbf{z}} \) will converge to the expected optimal value. The ADMM algorithm is typically used for solving convex problems; however, in machine learning, we often deal with non-convex optimization problems. Nevertheless, there are studies that demonstrate this algorithm can still converge to a local solution under non-convex conditions \cite{wang2018globalconvergenceadmmnonconvex,magnússon2015convergencealternatingdirectionlagrangian,YANG2022110551,yuan2024admmnonconvexoptimizationminimal}. For instance, the convergence of the ADMM algorithm for non-convex and structured optimization problems has been investigated, demonstrating that under certain conditions, ADMM can converge to a stationary point that satisfies the first-order necessary conditions.
The presented analysis relies on assumptions such as differentiability and regularity, making ADMM applicable to a broad range of non-convex problems in distributed optimization settings \cite{magnússon2015convergencealternatingdirectionlagrangian}. In the chapter \ref{Experimental Results}, our experimental tests also indicate that applying this optimization method to a non-convex objective function yields satisfactory performance.

\begin{figure}
    \centering
    \includegraphics[width=0.6\linewidth]{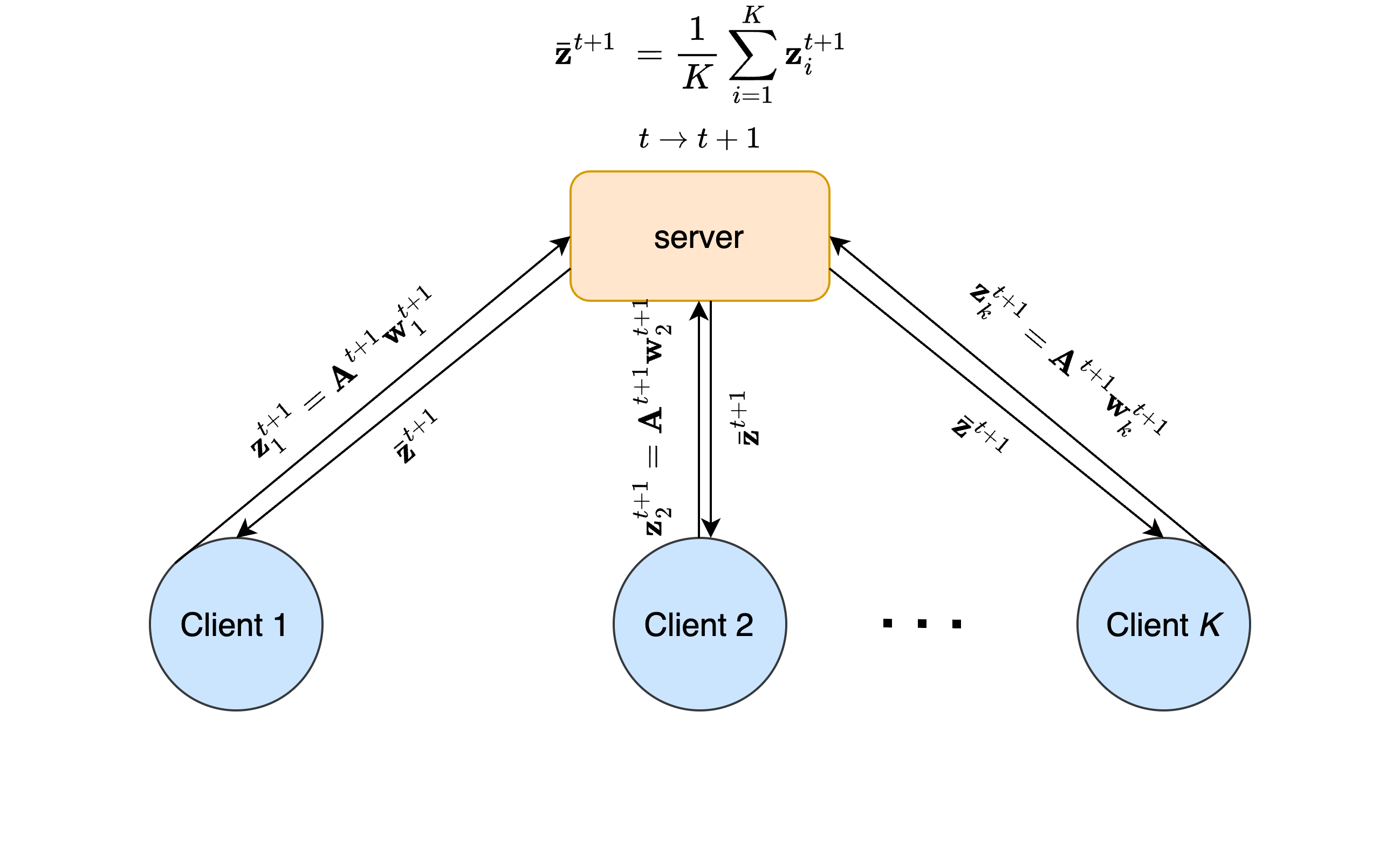}
    \caption{
        Structure of FedRP. In each round \( t \), each client first updates its local model \( \mathbf{w}_i^{t+1} \). Then, a new random projection matrix \( \mathbf{A}^{t+1} \) is generated and multiplied by \( \mathbf{w}_i^{t+1} \) to compute the compressed vector \( \mathbf{z}_i^{t+1}\). These vectors are sent to the server, which aggregates them as \( \bar{\mathbf{z}}^{t+1}\) and returns the result to the clients. This process repeats until convergence.
    }
    \label{fig:enter-label}
\end{figure}

\subsection{FedRP}

Federated learning struggles with privacy and communication costs. We propose a solution using random projection and the ADMM algorithm that maintains model performance while improving privacy and reducing communication costs. Random projection is a dimensionality reduction technique that maps high-dimensional data to a lower-dimensional space using a random matrix. This method leverages the Johnson-Lindenstrauss lemma, which guarantees that pairwise distances between points are approximately preserved in the lower-dimensional space. This technique is computationally efficient, thus accelerating algorithms and reducing storage requirements while preserving the data structure. In our method, during each training round, we generate a matrix whose elements are independently sampled from a normal distribution, and this matrix is shared among all clients. Random projection is then applied to the model parameters of each client before being sent to the central server. In other words, the randomization process is repeated in each training round, and using the ADMM algorithm, we achieve consensus among the clients’ model parameters in the new reduced space \cite{freksen2021introductionjohnsonlindenstrausstransforms,article,nabil2017randomprojectionapplications}.

In each training round, we define the projected (i.e., dimensionally reduced) model parameters sent to the server as vector \( \mathbf{z} \), where \( \mathbf{z} = A \mathbf{w} \). To obtain this vector, we create a matrix \( A \) with dimensions \( m \times n \), where \( m \) represents the dimension of the new parameter space and \( n \) represents the number of model parameters. Each element of the matrix is drawn independently of a normal distribution \( \mathcal{N}\left(0, \frac{1}{m}\right) \). 
All clients generate the same projection matrix using a shared seed unknown to the server, ensuring that the server cannot reconstruct the matrix. To mitigate computational instability and prevent the potential growth of vector \(\mathbf{z}\) values, we avoid generating the elements of matrix \( A \) from a standard normal distribution. In high-dimensional models, a standard normal distribution can lead to imbalanced values and numerical issues. Instead, we use a zero-mean normal distribution with a variance of \(\frac{1}{m}\). This adjustment counteracts the challenges posed by high-dimensional parameters, ensuring that the elements of matrix \( A \) are appropriately scaled. With this scaling, the expected squared norm of the projected vector equals that of the original parameter vector (\(\mathbb{E}[\|\mathbf{A}\mathbf{w}\|^2] = \|\mathbf{w}\|^2\)), preventing uncontrolled growth or shrinkage and keeping the values of \(\mathbf{z}\) in a stable range for reliable computation.

To securely generate the matrix A without granting the central server access to it, a secure sharing mechanism can be employed. In this approach, during each training round, one of the clients generates the matrix A using a specific seed that is different from those used in previous rounds. This seed is then encrypted using the public keys of the other clients and sent to them via the central server. Upon receiving and decrypting the seed, the clients independently generate the matrix A. This method ensures that all clients generate the same matrix without revealing it to the central server. An alternative would be to encrypt and transmit matrix \( A \) directly through the server to the other clients, which would entail higher communication costs due to the data volume. Employing cryptographic methods not only reduces communication costs, but also enhances security and privacy. In general, in FedRP, the central server does not require access to the matrix \( A \). This feature not only simplifies computational operations, but also enhances privacy, which will be explored theoretically in the next section.

As illustrated in Figure \ref{fig:enter-label}, each client multiplies its model parameters, represented by an n-dimensional vector \(\mathbf{w}_i\), by the matrix \( A \) before sending it to the central server. The resulting m-dimensional vector \( \mathbf{z}_i \) is then transmitted to the central server:
\begin{equation}
	\mathbf{z}_i = A\mathbf{w}_i. 
\end{equation}

The central server then averages the received vectors \( \mathbf{z}_i \) and sends the aggregated vector \( \bar{\mathbf{z}} \) back to the clients.

Therefore, by rewriting Equation \ref{eq3} to reflect the new problem conditions and the shared vector $\bar{\mathbf{z}}$, the following loss function for client~$i$ is obtained:

\begin{equation}\label{eq9}
L_i(\mathbf{w}_i, \bar{\mathbf{z}}, \mathbf{y}_i) =  f_i(\mathbf{w}_i) + \mathbf{y}_i^T (A \mathbf{w}_i - \bar{\mathbf{z}}) + \frac{\rho}{2} \| A \mathbf{w}_i - \bar{\mathbf{z}} \|_2^2. 
\end{equation}

By minimizing the loss function above, each client~$i$ aims to find an optimized vector $\mathbf{w}_i$ such that its projection is close to the shared vector $\bar{\mathbf{z}}$, while simultaneously minimizing the local loss on its own dataset. Based on this loss function, the steps of the ADMM algorithm in our problem setting are updated as follows:

1.Update \(\mathbf{w}_i\) (Local Update):

\begin{equation}
\label{rr}
\mathbf{w}_i^{t+1}\ = \arg \min_{\mathbf{w}_i} \left( f_i(\mathbf{w}_i) + (\mathbf{y}_i^t)^T(\mathbf{A}^t \mathbf{w}_i - \bar{\mathbf{z}}^t) + \frac{\rho}{2} \| \mathbf{A}^t \mathbf{w}_i - \bar{\mathbf{z}}^t \|_2^2 \right)
\end{equation}

2.Update \(\mathbf{y}_i\) (Dual Variable Update):

The vector $\mathbf{y} \in \mathbb{R}^{m}$, representing the dual variable, has the same dimensions as the shared vector $\bar{\mathbf{z}}$. It is updated at iteration~$t$ as follows:
\begin{equation}
\mathbf{y}_i^{t+1}\ = \mathbf{y}_i^t + \rho \left(\mathbf{A}^t \mathbf{w}_i^{t+1} - \bar{\mathbf{z}}^{t} \right)
\end{equation}

3.Random generation of the projection matrix $A^{t+1}$ and update of $\mathbf{z}_i^{t+1}$:

\begin{equation}
\mathbf{z}_i^{t+1} = \mathbf{A}^{t+1} \mathbf{w}_i^{t+1}
\end{equation}

4. Update the shared vector \(\bar{\mathbf{z}}\) (Global update at the server):

\begin{equation}
\bar{\mathbf{z}}^{t+1}\  = \frac{1}{K} \sum_{i=1}^K  \mathbf{z}_i^{t+1} 
\end{equation}

As illustrated in Algorithm \ref{alg:p_method}, the above steps are repeated until the variable $\bar{\mathbf{z}}$ converges to its optimal value. Upon convergence, the projected parameters of all clients (i.e., $A\mathbf{w}_i$) become approximately equal to the global vector $\bar{\mathbf{z}}$. Since the projected parameters are aligned, the probabilistic guarantee stated in Section~\ref{sec:projection_alignment} ensures that, with high probability, the original model parameters \(\mathbf{w}_i\) are also close to one another. Therefore, the model parameters from a single client can be used as the final global model accessible to the server.

\subsection{Projection Alignment Implies Parameter Alignment: A Probabilistic Guarantee}
\label{sec:projection_alignment}

When the projection dimension \(m\) is smaller than the model dimension \(n\), the mapping \(\mathbf{w} \mapsto \mathbf{A}\mathbf{w}\) is many‑to‑one.  Therefore, observing that all projected vectors \(\mathbf{z}_i = \mathbf{A}\mathbf{w}_i\) are close to each other does not automatically prove that the original weight vectors \(\mathbf{w}_i\) are close.  We now show that due to the randomness of \(\mathbf{A}\), the probability that two genuinely different weight vectors produce very similar projections is exponentially small in \(m\).  Consequently, if the server observes tight clustering of the projected vectors, it can conclude -- with overwhelming confidence -- that the original models are also tightly clustered, and any of them can serve as the global model.

\begin{theorem}[Projected proximity implies parameter proximity]
	\label{thm:projalign}
	Let $A\in\mathbb{R}^{m\times n}$ be a random projection matrix whose
	entries are independent and distributed as $\mathcal{N}(0,1/m)$, and let
	$\mathbf{w}_i,\mathbf{w}_j\in\mathbb{R}^n$ be any two fixed vectors. For every
	$\epsilon\in(0,1)$ and $\delta>0$, with probability at least
	\[
	1-\exp\!\left(-\frac{m\epsilon^2}{4}\right)
	\]
	over the draw of $A$, the following implication holds:
	\[
	\lVert A\mathbf{w}_i-A\mathbf{w}_j\rVert_2\leq\delta
	\quad\Longrightarrow\quad
	\lVert \mathbf{w}_i-\mathbf{w}_j\rVert_2
	\leq \frac{\delta}{\sqrt{1-\epsilon}}.
	\]
\end{theorem}

Theorem~\ref{thm:projalign} shows that, for any fixed pair of
client models, a small distance between their projected parameters is a
high-probability certificate that their original parameters are also close. Its proof which relies on a concentration lemma for the chi-squared distribution, is deferred to ~\ref{app:projalign}.

We can interpret that with overwhelming probability (exponentially close to 1 in \(m\)), if two projected vectors appear close, the original weight vectors are forced to be close up to a factor \(1/\sqrt{1-\varepsilon}\).  Because this bound is independent of the model size \(n\), it holds uniformly for any number of parameters, making the guarantee extremely robust.  The tightness of the bound improves as \(\varepsilon\) is chosen smaller, at a modest cost in the confidence level.

\begin{algorithm}
\caption{FedRP}\label{alg:p_method}
\textbf{Require:} The $K$ clients are indexed by $i$; $E$ is the number of local epochs, $A^0$ is initialize random projection matrix and $\rho > 0$ is the penalty rate.\\

\textbf{Server executes:}
\begin{algorithmic}[1]
    \STATE initialize $\bar{\mathbf{z}}^0$
    \FOR{each round $t=1,2,\dots$}
        \FOR{each client $i$ in parallel}
            \STATE ${\mathbf{z}}_i^{t+1} \leftarrow \text{ClientUpdate}(i, \bar{\mathbf{z}}^t)$
        \ENDFOR
        \STATE $\bar{\mathbf{z}}^{t+1} \leftarrow \frac{1}{K} \sum_{i=1}^K {\mathbf{z}}_i^{t+1}$
    \ENDFOR
\end{algorithmic}

\vspace{1em} 

\textbf{ClientUpdate} $(i, \bar{\mathbf{z}}^t)$: \texttt{// run on client $i$}
\begin{algorithmic}[1]
    \STATE $\mathbf{w}_i^{t+1} = \arg \min_{\mathbf{w}_i}\left( f_i(\mathbf{w}_i) + (\mathbf{y}_i^t)^T(\mathbf{A}^t \mathbf{w}_i - \bar{\mathbf{z}}^t) + \frac{\rho}{2} \| \mathbf{A}^t \mathbf{w}_i - \bar{\mathbf{z}}^t \|_2^2 \right)$
    \STATE $\mathbf{y}_i^{t+1} = \mathbf{y}_i^t + \rho (\mathbf{A}^t \mathbf{w}_i^{t+1} - \bar{\mathbf{z}}^t)$
    \STATE \texttt{Randomly generate $A^{t+1}$}
    \STATE ${\mathbf{z}}_i^{t+1} = \mathbf{A}^{t+1} \mathbf{w}_i^{t+1}$
    \STATE \textbf{return} ${{\mathbf{z}}}_i^{t+1}$
\end{algorithmic}
\end{algorithm}

\subsection{Privacy Analysis}

Differential Privacy is the primary metric for assessing the privacy of a randomized algorithm or mechanism, ensuring that the outcome of a mechanism is not much influenced by any single record in a dataset. In other words, whether a particular individual's data are included or not, the results of the algorithm remain approximately the same. This property is crucial for protecting the privacy of training data, as it ensures that changes to a single record do not significantly alter the output. To achieve this level of privacy, algorithms must be structured in such a way that the probability of obtaining a specific result from the database remains constant. In other words, the result of the model is nearly independent of any single record \cite{dpbook}.

\textbf{Definition 1:}\textit{(Neighboring Databases) Databases \( D \in \mathcal{D} \) and \( D' \in \mathcal{D} \) over a domain \( \mathcal{D} \) are called neighboring databases if they differ in exactly one record.}

This concept is crucial for understanding differential privacy, as it compares the effect of including or excluding a single record on the algorithm's output.

\textbf{Definition 2:}\textit{(Differential Privacy) A randomized algorithm \( A \) is \((\epsilon, \delta)\)-differentially private if for all neighboring databases \( D \) and \( D' \), and for all sets of outputs \( O \), the following inequality holds:
\begin{equation}
\Pr[A(D) \in O] \leq \exp(\epsilon) \cdot \Pr[A(D') \in O] + \delta
\end{equation}
}
Here, \(\Pr[\cdot]\) denotes the probability of an event. \(\epsilon\) and \(\delta\) in [0, 1] are used as quantitative measures to assess the level of privacy. The lower the value of \(\epsilon\), the better the privacy of our algorithm is preserved. When \(\delta = 0\), the model offers pure differential privacy. This means that the difference in the probabilities of any output between two neighboring databases is completely controlled by the factor \(\exp(\epsilon)\), and only the parameter \(\epsilon\) is used to determine the privacy level of the mechanism. In other words, the mechanism provides \(\epsilon\)-differential privacy. Furthermore, when \(\delta > 0\), it indicates that for some data in the dataset, our mechanism may not provide \(\epsilon\)-differential privacy, and the probability of this occurrence is denoted by \(\delta\). Although the parameter \(\delta\) partially relaxes the privacy guarantee compared to \(\epsilon\)-differential privacy, if \(\delta\) is negligible, it still ensures a significant level of protection against privacy breaches.

An important quantity in differential privacy is the $\ell_2$-sensitivity,
which measures the largest change, in $\ell_2$ norm, of a query or model
output when one data record is changed.  We define this quantity precisely
below for the model-training procedure considered here.

\textbf{Definition 3:}\textit{($\ell_2$-Sensitivity)}
Let $D$ and $D'$ be neighboring training datasets, and let $\mathbf{w}(D)$
denote the model parameters obtained by training on $D$.  The
$\ell_2$-sensitivity of the training procedure is
\[
\sup_{D\sim D'}\norm{\mathbf{w}(D)-\mathbf{w}(D')}.
\]
In the following theorem, this sensitivity is denoted by $\Delta$.  Thus,
$\Delta$ bounds the maximum change in the model parameters when the model is
trained on two neighboring datasets.

\begin{theorem}[Differential privacy of the FedRP mechanism]
	\label{ED}
	Consider the randomized mechanism
	\[
	\mathcal{M}(\mathbf{w})=A\mathbf{w},
	\]
	where, for each release, the entries of $A\in\mathbb{R}^{m\times n}$ are
	sampled independently according to
	\[
	A_{ij}\sim\mathcal{N}\!\left(0,\frac{1}{m}\right).
	\]
	Suppose that model vectors $\mathbf{w}$ and $\mathbf{w}'$ obtained from
	neighboring datasets satisfy
	\[
	\norm{\mathbf{w}-\mathbf{w}'}\leq\Delta,
	\qquad
	\norm{\mathbf{w}},\norm{\mathbf{w}'}\geq\sigma_{\min}>0.
	\]
	For any $\delta\in(0,1)$, define
	\begin{equation}
		\label{eq:parameters}
		\alpha:=\frac{\Delta}{\sigma_{\min}},
		\qquad
		\lambda_\delta:=\ln\frac{1}{\delta},
		\qquad
		T_\delta
		:=m+2\lambda_\delta
		+2\sqrt{\lambda_\delta(m+\lambda_\delta)}.
	\end{equation}
	If only $A\mathbf{w}$ is released, then $\mathcal{M}$ is
	$(\varepsilon,\delta)$-differentially private for
	\begin{equation}
		\label{EEEED}
		\varepsilon
		=\alpha\left(1+\frac{\alpha}{2}\right)T_\delta.
	\end{equation}
	Here, $m$ is the dimension of the projected space.
\end{theorem}

See \ref{appendix:A} for the proof. As can be seen from the theorem above, $\epsilon$ is proportional to the dimension $m$ as expected intuitively. The smaller $m$ is, the more privacy is preserved (smaller $\epsilon$).
An important parameter in this theorem is $\sigma_{\min}$, which is introduced to prevent the model parameter vector $\mathbf{w}$ from collapsing to the zero vector. If $\mathbf{w}$ becomes zero, then multiplying it by any matrix including the random projection matrix results in a zero vector. Consequently, the output of the algorithm becomes independent of the random matrix, and the mechanism loses its differential privacy guarantees. To avoid this, a lower bound $\sigma_{\min}$ is imposed. However, a smaller $\sigma_{\min}$ allows $\mathbf{w}$ to be closer to zero, which in turn increases the value of $\epsilon$ according to the theorem, thereby weakening the privacy protection.


It is worth noting that the above theorem quantifies the privacy guarantee $(\epsilon, \delta)$ for a single communication round between the server and clients. When the algorithm is executed for $T$ rounds, the overall privacy loss can be bounded using the advanced composition theorem for differential privacy~\cite{dwork2010boosting,kairouz2015composition}. In particular, for any $\delta' > 0$, the full sequence of $T$ mechanisms satisfies $(\epsilon_{\text{total}}, T\delta + \delta')$-differential privacy with $\epsilon_{\text{total}} = O\big(\epsilon \sqrt{T \log(1/\delta')}\big)$, which is significantly tighter than the naive linear bound $T\cdot\epsilon$ and allows for a much larger number of communication rounds under a fixed overall privacy budget. This bound can be further tightened by employing the Rényi Differential Privacy (RDP) accountant~\cite{mironov2017renyi}, which tracks the privacy loss moments across rounds and provides a sharper conversion from the RDP domain to the standard $(\epsilon,\delta)$-DP guarantee.

\section{Experimental Results}\label{Experimental Results}
\renewcommand{\arraystretch}{3} 
\setlength{\tabcolsep}{10pt} 

\begin{table*}[t]
	\centering
	\caption{Detailed architecture of the custom CNN model used for CIFAR10 experiments.}
	\label{tab:architecture-my-model}
	\footnotesize
	\setlength{\tabcolsep}{4pt}
	\renewcommand{\arraystretch}{0.85}   
	\begin{tabular}{@{}llcl@{}}
		\toprule
		\textbf{Block} & \textbf{Layer} & \textbf{Output Shape} & \textbf{Details} \\
		\midrule
		\multirow{5}{*}{Block 1}
		& Conv2d(3, 32, 3, padding=1)   & (32, $H$, $W$)     & kernel=3$\times$3, pad=1 \\
		& GroupNorm(8, 32) + ReLU        & (32, $H$, $W$)     & 8 groups \\
		& Conv2d(32, 32, 3, padding=1)  & (32, $H$, $W$)     & kernel=3$\times$3, pad=1 \\
		& GroupNorm(8, 32) + ReLU        & (32, $H$, $W$)     & 8 groups \\
		& MaxPool2d(2, 2)                & (32, $H$/2, $W$/2) & stride=2 \\
		\midrule
		\multirow{5}{*}{Block 2}
		& Conv2d(32, 64, 3, padding=1)  & (64, $H$/2, $W$/2) & kernel=3$\times$3, pad=1 \\
		& GroupNorm(16, 64) + ReLU       & (64, $H$/2, $W$/2) & 16 groups \\
		& Conv2d(64, 64, 3, padding=1)  & (64, $H$/2, $W$/2) & kernel=3$\times$3, pad=1 \\
		& GroupNorm(16, 64) + ReLU       & (64, $H$/2, $W$/2) & 16 groups \\
		& MaxPool2d(2, 2)                & (64, $H$/4, $W$/4) & stride=2 \\
		\midrule
		\multirow{5}{*}{Block 3}
		& Conv2d(64, 128, 3, padding=1) & (128, $H$/4, $W$/4) & kernel=3$\times$3, pad=1 \\
		& GroupNorm(16, 128) + ReLU      & (128, $H$/4, $W$/4) & 16 groups \\
		& Conv2d(128, 128, 3, padding=1)& (128, $H$/4, $W$/4) & kernel=3$\times$3, pad=1 \\
		& GroupNorm(16, 128) + ReLU      & (128, $H$/4, $W$/4) & 16 groups \\
		& MaxPool2d(2, 2)                & (128, $H$/8, $W$/8)& stride=2 \\
		\midrule
		\multirow{3}{*}{FC Layers}
		& Linear(128\,$\times$\,$HW$/64, 256) & (256)   & flatten + fc1 \\
		& Linear(256, 128)                & (128)   & fc2 \\
		& Linear(128, $C$)                & ($C$)   & fc3 \\
		\bottomrule
	\end{tabular}
	\renewcommand{\arraystretch}{1}   
\end{table*}


In this section, we evaluate FedRP algorithm using two datasets: MNIST and CIFAR-10, to compare its performance with conventional federated learning techniques. We examine the model's performance under varying conditions, such as different numbers of clients and random projection dimensions, to understand how these factors impact both accuracy and privacy. Specifically, we evaluate FedRP algorithm in three key areas: model performance and accuracy, privacy preservation, and communication costs between clients and the central server. Furthermore, we present a sensitivity analysis experiment for the projection dimension to see its effect on accuracy and convergence.

\subsection{Model Structure}

The evaluation utilizes well-known deep learning model, LeNet-5 \cite{lenet} for MNIST dataset, a custom CNN model for CIFAR10 dataset, which vary in complexity and size, to test the algorithm under different conditions. LeNet-5 is a simple convolutional neural network (CNN) with approximately 60,000 parameters, serving as a base model. The custom CNN is a network architecture consisting of 9 trainable layers and approximately 800,000 parameters. The architecture of this model is presented in Table~\ref{tab:architecture-my-model}. We avoided larger models like Resnet18 and VGG16 because we couldn't test them with large enough projection dimensions due to hardware constraints.
Each model was trained with a batch size of 64, and evaluated with a batch size of 16.

\subsection{Dataset}

As mentioned, we use two standard deep learning datasets: MNIST \cite{mnist} and CIFAR-10 \cite{cifar}. MNIST includes 70,000 grayscale images of handwritten digits (0-9) at 28x28 resolution. CIFAR-10 consists of 60,000 color images (32x32) across 10 object categories.

\subsection{Federated Learning Setup}

The federated learning environment is designed according to the complexity of the model. For LeNet-5, we tested with 100 clients. For the custom CNN, the number of clients was set to 50. We reduced the number of clients for the custom CNN which is more complex than Lenet-5 due to processing limitations and available memory. The models were trained using different federated learning algorithms, including FedAvg, FedADMM, FedAvg+DP, FWC and FedRP.
\subsubsection{Algorithms Compared}
\begin{itemize}
    \item \textbf{FedAvg:} A foundational and widely used federated learning algorithm that aggregates local model parameters by averaging. We use the basic version of this algorithm \cite{mcmahan2023communicationefficientlearningdeepnetworks}.
    
    \item \textbf{FedADMM:} A distributed optimization technique described in the section \ref{admmmmm} \cite{gong2022fedadmmrobustfederateddeep}.
   
    \item \textbf{FedAvg+DP:} An enhanced version of FedAvg that incorporates privacy-preserving techniques such as differential privacy. Gaussian noise was added to the model parameters before sending them to the central server \cite{mcmahan2019generalapproachaddingdifferential}  to achieve different levels of privacy. We tried to get the best privacy level possible while the training does not fail completely.
    
    \item \textbf{Federated without Connection (FWC):} A version of federated learning where clients operate independently without shared information. We used this algorithm in our evaluations to determine the minimum achievable accuracy for different models. We report the average accuracy of clients based on their performance on the global test set.

    \item \textbf{FedRP:} Our algorithm utilizes random projection to reduce the dimension of the model, with mapping dimensions tuned according to the memory and processing capabilities.
    
    In the test phase, the central server can use either a randomly selected client model or the average of all client models.
\end{itemize}
All training was performed with balanced data splits between clients. Each client trained for 1 and 5 local epochs in their entire train dataset per round, allowing for detailed performance evaluation.

\begin{table*}[t]
    \centering
    \renewcommand{\arraystretch}{1.2} 
    \setlength{\tabcolsep}{6pt} 
    \caption{The amount of information sent to the server from each client in each communication round.}
    \label{tab:info_sent}
    \adjustbox{max width=\textwidth}{ 
    \begin{tabular}{l c c c c c c c c}
        \toprule
        \textbf{Model} & \textbf{FedAvg} & \multicolumn{7}{c}{\textbf{Proposed Algorithm}} \\
        \cmidrule(lr){3-9}
         &  & RPD=1 & RPD=5 & RPD=10 & RPD=50 & RPD=100 & RPD=1000 & RPD=10000 \\
        \midrule
        LeNet-5  & 241 KB  & 4 Byte  & 20 Byte  & 40 Byte  & 200 Byte  & 400 Byte  & 4 KB  & 40 KB  \\
        ResNet-18  & 42 MB  & 4 Byte  & 20 Byte  & 40 Byte  & 200 Byte  & 400 Byte  & 4 KB  & 40 KB  \\
        VGG-16  & 512 MB  & 4 Byte  & 20 Byte  & 40 Byte  & 200 Byte  & 400 Byte  & 4 KB  & 40 KB  \\
        \bottomrule
    \end{tabular}
    }
\end{table*}

\subsection{Accuracy and Performance}

Figure \ref{fig:mainfig} and \ref{fig:mainfig2} show the performance of FedRP compared to other algorithms on various datasets, with 1 and 5 local epochs, respectively. In all of these graphs, the performance of FedRP is better than that of the FedAvg+DP algorithm, which is recognized as one of the primary conventional algorithms in the field of privacy preservation. Furthermore, in most cases, there is only a slight difference in accuracy and performance compared to the FedAvg algorithm.

The results demonstrate that FedRP exhibits satisfactory performance in terms of accuracy and efficiency, thereby effectively addressing the primary challenges inherent to privacy-preserving algorithms, particularly concerning potential declines in performance and accuracy. Additionally, it is noteworthy that the dimensionality of the random projection matrix did not exert a significant impact on performance. Consequently, opting for a reduced dimensionality could prove to be an optimal strategy, as it augments privacy preservation while substantially diminishing communication costs.

\begin{figure*}[ht]
	\pgfplotscreateplotcyclelist{mycolors}{%
		blue,thick,mark=*,mark size=1.0\\%
		red,thick,mark=*,mark size=1.0\\%
		orange,thick,mark=*,mark size=1.0\\%
		purple,thick,mark=*,mark size=1.0\\%
		cyan,thick,mark=*,mark size=1.0\\%
		green!50!black,thick,mark=*,mark size=1.0\\%
	}
	\centering
	\ref*{globallegend}\\[2mm]
	
	\subfigure[LeNet5 on MNIST]{%
		\begin{tikzpicture}
			\begin{axis}[
				width=0.49\linewidth,
				height=0.42\linewidth,
				xlabel={Round},
				ylabel={Test Accuracy (\%)},
				ymin=20, ymax=100,
				legend to name=globallegend,
				legend style={
					font=\scriptsize,
					legend columns=3,
					draw=none
				},
				grid=major,
				cycle list name=mycolors,
				table/col sep=comma
				]
				\addplot+ table [x index=0, y index=1] {hh/lenet5_experiments_sgd.csv};
				\addlegendentry{FedRP}
				\addplot+ table [x index=0, y index=2] {hh/lenet5_experiments_sgd.csv};
				\addlegendentry{FedADMM}
				\addplot+ table [x index=0, y index=3] {hh/lenet5_experiments_sgd.csv};
				\addlegendentry{FedAvg}
				\addplot+ table [x index=0, y index=4] {hh/lenet5_experiments_sgd.csv};
				\addlegendentry{FedAvgDP($\epsilon=10$)}
				\addplot+ table [x index=0, y index=5] {hh/lenet5_experiments_sgd.csv};
				\addlegendentry{FedAvgDP($\epsilon=5$)}
				
				\addplot [dashed, black, thick] coordinates {
					(1, 88.0)   
					(30, 88.0)
				};
				\addlegendentry{FWC}
				
				\addlegendimage{line legend, green!50!black, thick, mark=*, mark size=1.0}
				\addlegendentry{FedAvgDP($\epsilon=40$)}
			\end{axis}
		\end{tikzpicture}
		\label{fig:subfig1}
	}%
	\hspace{-2mm}%
	\subfigure[Custom CNN on CIFAR10]{%
		\begin{tikzpicture}
			\begin{axis}[
				width=0.49\linewidth,
				height=0.42\linewidth,
				xlabel={Round},
				ymin=10, ymax=80,
				grid=major,
				cycle list name=mycolors,
				table/col sep=comma,
				legend style={
					font=\scriptsize,
					at={(0.98,0.98)},
					anchor=north east,
					draw=none
				}
				]
				\addplot+ table [x index=0, y index=1] {hh/custom_cifar10_experiments.csv};
				\addplot+ table [x index=0, y index=2] {hh/custom_cifar10_experiments.csv};
				\addplot+ table [x index=0, y index=3] {hh/custom_cifar10_experiments.csv};
				\addplot+[color=green!50!black] table [x index=0, y index=4] {hh/custom_cifar10_experiments.csv};
				
				\addplot [dashed, black, thick] coordinates {
					(1, 35.0)   
					(40, 35.0)
				};
			\end{axis}
		\end{tikzpicture}
		\label{fig:subfig2}
	}
	\caption{Comparison of FedRP Algorithm Performance with Other Algorithms Under Different Conditions. (local epochs = 1)}
	\label{fig:mainfig}
\end{figure*}

\begin{figure*}[ht]
	\pgfplotscreateplotcyclelist{mycolors}{%
		blue,thick,mark=*,mark size=1.0\\%
		red,thick,mark=*,mark size=1.0\\%
		orange,thick,mark=*,mark size=1.0\\%
		purple,thick,mark=*,mark size=1.0\\%
		cyan,thick,mark=*,mark size=1.0\\%
		green!50!black,thick,mark=*,mark size=1.0\\%
	}
	\centering
	\ref*{globallegendE5}\\[2mm]   
	
	\subfigure[LeNet5 on MNIST]{%
		\begin{tikzpicture}
			\begin{axis}[
				width=0.49\linewidth,
				height=0.42\linewidth,
				xlabel={Round},
				ylabel={Test Accuracy (\%)},
				ymin=9, ymax=100,
				legend to name=globallegendE5,   
				legend style={
					font=\scriptsize,
					legend columns=4,
					draw=none,
					/tikz/every even column/.append style={column sep=0.3cm}
				},
				grid=major,
				cycle list name=mycolors,
				table/col sep=comma
				]
				\addplot+ table [x index=0, y index=1] {hh/mnist_experiments_E5.csv};
				\addlegendentry{FedRP}
				\addplot+ table [x index=0, y index=2] {hh/mnist_experiments_E5.csv};
				\addlegendentry{FedADMM}
				\addplot+ table [x index=0, y index=3] {hh/mnist_experiments_E5.csv};
				\addlegendentry{FedAvg}
				\addplot+ table [x index=0, y index=4] {hh/mnist_experiments_E5.csv};
				\addlegendentry{FedAvgDP($\epsilon=5$)}
				\addplot+ table [x index=0, y index=5] {hh/mnist_experiments_E5.csv};
				\addlegendentry{FedAvgDP($\epsilon=10$)}
				
				\addplot [dashed, black, thick] coordinates {
					(1, 88.0)
					(30, 88.0)
				};
				\addlegendentry{FWC}
				
				\addlegendimage{line legend, green!50!black, thick, mark=*, mark size=1.0}
				\addlegendentry{FedAvgDP(1,3)}          
				
				\addlegendimage{line legend, cyan!50!black, thick, mark=*, mark size=1.0}
				\addlegendentry{FedAvgDP($\epsilon=10$)}         
				
				\addlegendimage{line legend, green!70!black, thick, mark=*, mark size=1.0}
				\addlegendentry{FedAvgDP($\epsilon=30$)}         
			\end{axis}
		\end{tikzpicture}
		\label{fig:subfig1}
	}%
	\hspace{-2mm}%
	\subfigure[Custom CNN on CIFAR10]{%
		\begin{tikzpicture}
			\begin{axis}[
				width=0.49\linewidth,
				height=0.42\linewidth,
				xlabel={Round},
				ymin=10, ymax=80,
				grid=major,
				cycle list name=mycolors,
				table/col sep=comma,
				legend style={
					font=\scriptsize,
					at={(0.98,0.98)},
					anchor=north east,
					draw=none
				}
				]
				\addplot+ table [x index=0, y index=1] {hh/cifar10_experiments_E5.csv};
				\addplot+ table [x index=0, y index=2] {hh/cifar10_experiments_E5.csv};
				\addplot+ table [x index=0, y index=3] {hh/cifar10_experiments_E5.csv};
				\addplot+[color=green!50!black] table [x index=0, y index=4] {hh/cifar10_experiments_E5.csv};
				\addplot+[color=cyan!50!black] table [x index=0, y index=5] {hh/cifar10_experiments_E5.csv};
				
				\addplot [dashed, black, thick] coordinates {
					(1, 35.0)
					(40, 35.0)
				};
			\end{axis}
		\end{tikzpicture}
		\label{fig:subfig2}
	}
	\caption{Comparison of FedRP Algorithm Performance with Other Algorithms Under Different Conditions. (local epochs = 5)}
	\label{fig:mainfig2}
\end{figure*}

\begin{figure*}[ht]
	\centering
	\ref*{globallegendSens}\\[2mm]
	
	\subfigure[LeNet5 on MNIST]{%
		\begin{tikzpicture}
			\begin{axis}[
				width=0.49\linewidth,
				height=0.42\linewidth,
				xlabel={Round},
				ylabel={Test Accuracy (\%)},
				ymin=10, ymax=100,
				legend to name=globallegendSens,      
				legend style={
					font=\scriptsize,
					legend columns=3,
					draw=none
				},
				grid=major,
				table/col sep=comma
				]
				\addlegendimage{line legend, blue,   thick, mark=*, mark size=1}
				\addlegendentry{$m=50000$}
				\addlegendimage{line legend, red,    thick, mark=*, mark size=1}
				\addlegendentry{$m=10000$}
				\addlegendimage{line legend, orange, thick, mark=*, mark size=1}
				\addlegendentry{$m=5000$}
				
				\addplot+[color=purple,         thick, mark=*, mark size=1] table [x index=0, y index=1] {hh/lenet5_ablation.csv};
				\addlegendentry{$m=1000$}
				\addplot+[color=cyan,           thick, mark=*, mark size=1] table [x index=0, y index=2] {hh/lenet5_ablation.csv};
				\addlegendentry{$m=500$}
				\addplot+[color=green!50!black, thick, mark=*, mark size=1] table [x index=0, y index=3] {hh/lenet5_ablation.csv};
				\addlegendentry{$m=100$}
				\addplot+[color=teal,           thick, mark=*, mark size=1] table [x index=0, y index=4] {hh/lenet5_ablation.csv};
				\addlegendentry{$m=50$}
				\addplot+[color=brown,          thick, mark=*, mark size=1] table [x index=0, y index=5] {hh/lenet5_ablation.csv};
				\addlegendentry{$m=10$}
				\addplot+[color=pink,           thick, mark=*, mark size=1] table [x index=0, y index=6] {hh/lenet5_ablation.csv};
				\addlegendentry{$m=1$}
			\end{axis}
		\end{tikzpicture}
		\label{fig:ablation_mnist}
	}%
	\hspace{5mm}%
	\subfigure[Custom CNN on CIFAR10]{%
		\begin{tikzpicture}
			\begin{axis}[
				width=0.49\linewidth,
				height=0.42\linewidth,
				xlabel={Round},
				ymin=10, ymax=60,
				grid=major,
				table/col sep=comma
				]
				\addplot+[color=blue,           thick, mark=*, mark size=1] table [x index=0, y index=1] {hh/cifar10_ablation.csv};
				\addplot+[color=red,            thick, mark=*, mark size=1] table [x index=0, y index=2] {hh/cifar10_ablation.csv};
				\addplot+[color=orange,         thick, mark=*, mark size=1] table [x index=0, y index=3] {hh/cifar10_ablation.csv};
				\addplot+[color=purple,         thick, mark=*, mark size=1] table [x index=0, y index=4] {hh/cifar10_ablation.csv};
				\addplot+[color=cyan,           thick, mark=*, mark size=1] table [x index=0, y index=5] {hh/cifar10_ablation.csv};
				\addplot+[color=green!50!black, thick, mark=*, mark size=1] table [x index=0, y index=6] {hh/cifar10_ablation.csv};
				\addplot+[color=brown,          thick, mark=*, mark size=1] table [x index=0, y index=7] {hh/cifar10_ablation.csv};
			\end{axis}
		\end{tikzpicture}
		\label{fig:ablation_cifar10}
	}
	\caption{Sensitivity of FedRP to the projection dimension $m$.}
	\label{fig:ablation}
\end{figure*}

\subsection{Sensitivity analysis}
The effect of the projection dimension 
$m$ on final global test accuracy is shown in Figure~\ref{fig:ablation}. This outcome aligns with theoretical expectations: in a drastically reduced projection space, two sets of original weights can be far apart even when their projected counterparts appear close, i.e., the projection no longer preserves distances with high probability.
\subsection{Final Word on Privacy Preservation}

The privacy of the proposed algorithm can be analyzed from two perspectives: first, in terms of the values of $(\epsilon, \delta)$-DP that the algorithm guarantees, and second, its performance against data reconstruction attacks, such as the DLG attack. We can not compare the $\epsilon$ values of the two methods because we can't calculate $\Delta$ for FedRP. In other words, we can't find an upper bound for the difference of the model parameters trained on neighbouring datasets.

About the second method, we mentioned, most user data reconstruction methods aimed at privacy threats and attacks require access to gradients sent from clients to the central server or to two consecutive versions of model parameters sent by clients to the server \cite{DLG}. In FedRP algorithm, both of these conditions are not met. In this approach, instead of sending the full model parameters, clients send only a dimension-reduced vector of these parameters to the server. Due to the structure of the random mapping matrix, it is impossible to accurately reconstruct this matrix and the original model parameters from the reduced-dimension vector, as there is no analytical and error-free solution for estimating these values. Thus, FedRP can effectively resist many common data reconstruction attacks.

\subsection{Communication Cost}

As you can see in Figure \ref{fig:mainfig}, the range of variations in FedRP's performance under different dimensions of the random mapping matrix is low and it is observed that the precision of FedRP, under the same number of communication rounds, is very close to that of the algorithm FedAvg. However, there is a fundamental difference between the two algorithms. In the FedAvg algorithm, clients send the complete model parameters to the central server in each communication round. For instance, in the case of the VGG16 model, these parameters form a vector of approximately 138 million dimensions. In contrast, in FedRP, each client only sends a vector with chosen dimensions, which could be, for example, of dimension 1 or 10, to the central server.

This feature significantly reduces the volume of communication data between clients and the central server. In other words, the lower the dimension of the random projection, the lower the communication cost. This reduction in communication cost, while maintaining model accuracy, highlights the high efficiency of FedRP compared to traditional methods such as FedAvg, resulting in a significant improvement in the preservation of privacy and the reduction of communication costs. In Table \ref{tab:info_sent}, you can see the comparison of the communication cost of FedRP with the FedAvg algorithm. As shown in Table \ref{tab:info_sent}, the volume of messages exchanged between clients and the central server in FedRP with a random projection dimension of 1 has decreased significantly compared to the FedAvg algorithm for the LeNet-5, ResNet-18 and VGG-16 models, with reductions of \(1.6 \times 10^{-5}\), \(9.5 \times 10^{-8}\), and \(7.8 \times 10^{-9}\), respectively.

As you observed in the results above, FedRP algorithm has achieved a proper trade-off between performance, privacy preservation, and communication cost. It outperforms the conventional DP algorithm in terms of performance while significantly surpassing the FedAvg algorithm in privacy preservation and communication cost.

\section{Conclusion}

Preserving user privacy in federated learning often forces a compromise between accuracy and communication cost. This paper introduced FedRP, a novel algorithm that combines random projection with the ADMM framework to jointly address both concerns. By projecting high-dimensional model parameters into a low-dimensional space before transmission, FedRP creates a one-way information bottleneck that prevents the server or adversaries from reconstructing the original data. A rigorous $(\epsilon,\delta)$-differential privacy guarantee was established, and the method was shown to be resilient against common reconstruction attacks such as DLG, because the server never accesses the raw model parameters.

Experiments on MNIST and CIFAR‑10 showed that FedRP achieves test accuracy close to FedAvg while substantially outperforming conventional differential privacy baselines (FedAvg+DP). Because only a compressed $m$-dimensional vector is transmitted per round, communication costs are reduced by orders of magnitude relative to full-parameter exchange. Notably, performance remained robust even for very small $m$, demonstrating a favorable privacy-communication-accuracy trade-off.

Several limitations point to future work. First, the algorithm currently assumes full client participation in every round, which is realistic only for cross-silo settings; extending it to cross-device scenarios with client sampling, stragglers, and dropouts is needed. Second, the analysis was conducted under the IID data assumption; investigating the impact of statistical heterogeneity on projection-based consensus and designing adaptive or personalized variants is an important next step. Third, for very large models, generating a fresh random matrix $\mathbf{A}$ and multiplying it by the weight vector each round introduces non-negligible computational overhead; fast Johnson-Lindenstrauss transforms or structured random matrices could alleviate this. Finally, the current practice of selecting a single random client's model as the global model, although backed by the pairwise probabilistic guarantee, may be sensitive to outliers; a more robust aggregation strategy, such as securely averaging the original parameters after the last round, would improve reliability.

Addressing these limitations will broaden FedRP's applicability to large-scale, non-IID, and resource‑constrained federated learning tasks while preserving its strong privacy and communication efficiency.


\bibliographystyle{elsarticle-harv} 
\bibliography{name}

\newpage
\appendix
\onecolumn

\section{Proof of Theorem~\ref{thm:projalign} (Projection Alignment)}
\label{app:projalign}

The proof relies on the following concentration lemma. The goal of the following lemma is to control how much the random projection
can shrink the distance between two fixed vectors. In particular, it shows
that the squared projected distance
\(\|\mathbf{A}\mathbf{w}_i-\mathbf{A}\mathbf{w}_j\|_2^2\) is unlikely to fall
far below the original squared distance
\(\|\mathbf{w}_i-\mathbf{w}_j\|_2^2\).
\begin{lemma}[Lower-tail concentration of the projected distance]
	\label{lem:projconc}
	For any fixed difference vector \(\mathbf{v} = \mathbf{w}_i - \mathbf{w}_j\) with \(d = \|\mathbf{v}\|_2 > 0\) and any \(\varepsilon \in (0,1)\),
	\begin{equation}
		\Pr\!\Big[\|\mathbf{A}\mathbf{v}\|_2^2
		< (1-\varepsilon)d^2\Big]
		\;\le\; \exp\!\Bigl(-\frac{m\varepsilon^2}{4}\Bigr).
	\end{equation}
\end{lemma}

\begin{proof}[Proof of Lemma~\ref{lem:projconc}]
	Each row of \(\mathbf{A}\) is independent and has distribution \(\mathcal{N}(0,\frac{1}{m}\mathbf{I}_n)\).  For a fixed \(\mathbf{v}\), the vector \(\mathbf{A}\mathbf{v} \in \mathbb{R}^m\) has independent entries, each distributed as \(\mathcal{N}(0, d^2/m)\).  Thus
	\[
	\|\mathbf{A}\mathbf{v}\|_2^2 \;=\; \frac{d^2}{m} \sum_{k=1}^m X_k^2,
	\]
	where \(X_k \sim \mathcal{N}(0,1)\) are independent standard normal random variables.  The sum of squares \(Y = \sum_{k=1}^m X_k^2\) follows a chi-squared distribution with \(m\) degrees of freedom, \(\chi^2_m\). The Laurent--Massart lower-tail inequality~\cite[Lemma~1]{laurent2000adaptive} states that, for every \(x>0\),
	\[
	\Pr\!\left[Y-m\leq-2\sqrt{mx}\right]\leq e^{-x}.
	\]
	Taking \(x=m\varepsilon^2/4\) gives
	\[
	\Pr\!\Big[Y\leq(1-\varepsilon)m\Big]
	\leq \exp\!\Bigl(-\frac{m\varepsilon^2}{4}\Bigr).
	\]
	Using \(\|\mathbf{A}\mathbf{v}\|_2^2=(d^2/m)Y\) proves the lemma.
\end{proof}

\begin{proof}[Proof of Theorem~\ref{thm:projalign}]
	Let
	\[
	\mathbf{v}=\mathbf{w}_i-\mathbf{w}_j,
	\qquad
	d=\|\mathbf{v}\|_2.
	\]
	Then \(A\mathbf{v}=A\mathbf{w}_i-A\mathbf{w}_j\). If \(d=0\), the two
	original vectors are identical, so the conclusion holds immediately. We may
	therefore assume that \(d>0\).
	
	The implication in the theorem could fail only if the projection shrinks
	the difference vector too much. To exclude this possibility, consider the
	event
	\[
	\mathcal{E}
	=\left\{\|A\mathbf{v}\|_2^2\geq(1-\varepsilon)d^2\right\}.
	\]
	By Lemma~\ref{lem:projconc}, this event has probability at least
	\[
	\Pr[\mathcal{E}]
	\geq 1-\exp\!\left(-\frac{m\varepsilon^2}{4}\right).
	\]
	On \(\mathcal{E}\), taking square roots gives the lower bound
	\[
	\|A\mathbf{w}_i-A\mathbf{w}_j\|_2
	=\|A\mathbf{v}\|_2
	\geq\sqrt{1-\varepsilon}\,d.
	\]
	Now suppose that the projected distance is at most \(\delta\). Combining
	the preceding lower bound with this assumption yields
	\[
	\sqrt{1-\varepsilon}\,
	\|\mathbf{w}_i-\mathbf{w}_j\|_2
	\leq
	\|A\mathbf{w}_i-A\mathbf{w}_j\|_2
	\leq\delta.
	\]
	Since \(\varepsilon\in(0,1)\), division by
	\(\sqrt{1-\varepsilon}\) gives
	\[
	\|\mathbf{w}_i-\mathbf{w}_j\|_2
	\leq\frac{\delta}{\sqrt{1-\varepsilon}}.
	\]
	Therefore, the desired implication holds on \(\mathcal{E}\), which occurs
	with the probability stated in the theorem.
\end{proof}

\section{Proof of Theorem~\ref{ED} (Differential Privacy of FedRP)}\label{appendix:A}

We prove the result for one release of the projected vector.  The sampled
matrix $A$ is internal randomness and is not included in the released
output.

Write the rows of $A$ as
\[
A=
\begin{bmatrix}
	\mathbf{a}_1^\top\\
	\vdots\\
	\mathbf{a}_m^\top
\end{bmatrix},
\qquad
\mathbf{a}_j\sim\mathcal{N}\!\left(0,\frac{1}{m}I_n\right).
\]
The $j$th coordinate of the output is therefore
\[
(A\mathbf{w})_j=\mathbf{a}_j^\top\mathbf{w}
\sim\mathcal{N}\!\left(0,\frac{\norm{\mathbf{w}}^2}{m}\right).
\]
The rows of $A$ are independent, so
\[
\mathcal{M}(\mathbf{w})
\sim
\mathcal{N}\!\left(0,\frac{\norm{\mathbf{w}}^2}{m}I_m\right).
\]
In particular,
$\mathbb{E}\norm{\mathcal{M}(\mathbf{w})}^2=\norm{\mathbf{w}}^2$, which is the
usual normalization for a Gaussian random projection.

For distributions $P$ and $Q$ with densities $p$ and $q$, define the privacy
loss in the direction from $P$ to $Q$ by
\[
L_{P\|Q}(\mathbf{z}):=\ln\frac{p(\mathbf{z})}{q(\mathbf{z})}.
\]
If
\[
\Pr_{\mathbf{z}\sim P}
\!\left(L_{P\|Q}(\mathbf{z})>\varepsilon\right)\leq\delta,
\]
then $P(S)\leq e^\varepsilon Q(S)+\delta$ for every measurable set $S$.
Therefore, it suffices to bound the privacy loss in both directions between
the two neighboring output distributions.

\subsubsection*{Proof of Theorem~\ref{ED}}

\begin{proof}
	Define the smaller and larger input norms by
	\[
	r_-:=\min\!\left\{\norm{\mathbf{w}},\norm{\mathbf{w}'}\right\},
	\qquad
	r_+:=\max\!\left\{\norm{\mathbf{w}},\norm{\mathbf{w}'}\right\},
	\]
	and let
	\[
	\rho:=\frac{r_+}{r_-}\geq1.
	\]
	The reverse triangle inequality gives
	\[
	r_+-r_-
	\leq\norm{\mathbf{w}-\mathbf{w}'}
	\leq\Delta.
	\]
	Since $r_-\geq\sigma_{\min}$, we obtain
	\begin{equation}
		\label{eq:norm-ratio}
		1\leq\rho
		=1+\frac{r_+-r_-}{r_-}
		\leq1+\frac{\Delta}{\sigma_{\min}}
		=1+\alpha.
	\end{equation}
	
	The corresponding Gaussian densities are, for
	$\mathbf{z}\in\mathbb{R}^m$,
	\[
	p_\pm(\mathbf{z})
	=\left(\frac{m}{2\pi r_\pm^2}\right)^{m/2}
	\exp\!\left(-\frac{m\norm{\mathbf{z}}^2}{2r_\pm^2}\right).
	\]
	
	\medskip
	\noindent\textit{Direction 1: from the smaller-norm output distribution to
		the larger-norm output distribution.}
	For any fixed $\mathbf{z}\in\mathbb{R}^m$, the Gaussian density formula gives
	\begin{align*}
		L_{-\|+}(\mathbf{z})
		:=\ln\frac{p_-(\mathbf{z})}{p_+(\mathbf{z})}
		&=m\ln\rho
		-\frac{m\norm{\mathbf{z}}^2}{2}
		\left(\frac{1}{r_-^2}-\frac{1}{r_+^2}\right) \\
		&\leq m\ln\rho
		\leq m\ln(1+\alpha).
	\end{align*}
	Thus, in this direction, the privacy loss is bounded deterministically by
	\begin{equation}
		\label{eq:epsilon-minus-plus}
		\varepsilon_{-\|+}:=m\ln(1+\alpha).
	\end{equation}
	
	\medskip
	\noindent\textit{Direction 2: from the larger-norm output distribution to
		the smaller-norm output distribution.}
	For any fixed $\mathbf{z}\in\mathbb{R}^m$, the reverse privacy loss is
	\[
	L_{+\|-}(\mathbf{z})
	:=\ln\frac{p_+(\mathbf{z})}{p_-(\mathbf{z})}
	=-m\ln\rho
	+\frac{\rho^2-1}{2}
	\frac{m\norm{\mathbf{z}}^2}{r_+^2}.
	\]
	Now let
	\[
	\mathbf{z}_+
	\sim\mathcal{N}\!\left(0,\frac{r_+^2}{m}I_m\right)
	\]
	and define
	\[
	Y:=\frac{m\norm{\mathbf{z}_+}^2}{r_+^2}.
	\]
	Because $\sqrt{m}\,\mathbf{z}_+/r_+$ is a standard Gaussian vector in
	$\mathbb{R}^m$, we have $Y\sim\chi_m^2$.  Evaluating the preceding privacy-loss
	function at $\mathbf{z}_+$ gives
	\begin{align*}
		L_{+\|-}(\mathbf{z}_+)
		&=-m\ln\rho+\frac{\rho^2-1}{2}Y \\
		&\leq\frac{(1+\alpha)^2-1}{2}Y \\
		&=\alpha\left(1+\frac{\alpha}{2}\right)Y,
	\end{align*}
	where we used~\eqref{eq:norm-ratio} and discarded the nonpositive term
	$-m\ln\rho$.
	
	For $Y\sim\chi_m^2$, we use the upper-tail bound
	\[
	\Pr(Y-m\geq t)
	\leq
	\exp\!\left(-\frac{t^2}{4(t+m)}\right),
	\qquad t\geq0.
	\]
	This inequality is a convenient consequence of the Laurent--Massart
	chi-squared concentration bound~\cite[Lemma~1]{laurent2000adaptive}.
	Set
	\[
	t_\delta
	:=2\lambda_\delta
	+2\sqrt{\lambda_\delta(m+\lambda_\delta)}.
	\]
	This is the positive solution of
	\[
	\frac{t_\delta^2}{4(t_\delta+m)}=\lambda_\delta.
	\]
	Since $T_\delta=m+t_\delta$ and
	$e^{-\lambda_\delta}=\delta$, the tail bound gives
	\[
	\Pr(Y>T_\delta)\leq\delta.
	\]
	Therefore, except on an event of probability at most $\delta$,
	\[
	L_{+\|-}(\mathbf{z}_+)
	\leq
	\alpha\left(1+\frac{\alpha}{2}\right)T_\delta.
	\]
	Thus, in the reverse direction, the privacy loss is bounded, except with
	probability at most $\delta$, by
	\begin{equation}
		\label{eq:epsilon-plus-minus}
		\varepsilon_{+\|-}
		:=\alpha\left(1+\frac{\alpha}{2}\right)T_\delta.
	\end{equation}
	
	To obtain one privacy parameter that is valid in both directions, take
	\[
	\varepsilon
	=\max\!\left\{\varepsilon_{-\|+},\varepsilon_{+\|-}\right\}.
	\]
	Since $\alpha\geq0$,
	\[
	\ln(1+\alpha)
	\leq\alpha
	\leq\alpha\left(1+\frac{\alpha}{2}\right),
	\qquad
	m\leq T_\delta.
	\]
	Consequently,
	\[
	\varepsilon_{-\|+}
	\leq\varepsilon_{+\|-},
	\]
	and hence
	\[
	\varepsilon
	=\varepsilon_{+\|-}
	=\alpha\left(1+\frac{\alpha}{2}\right)T_\delta.
	\]
	The privacy-loss criterion therefore proves that $\mathcal{M}$ is
	$(\varepsilon,\delta)$-differentially private.
\end{proof}

\end{document}